\documentclass[letterpaper]{article} 
\usepackage[preprint]{aaai2027}  
\usepackage[hyphens]{url}  
\usepackage{graphicx} 
\usepackage{natbib}  
\usepackage{caption} 
\usepackage{algorithm}
\usepackage{algorithmic}

\usepackage{newfloat}
\usepackage{listings}
\DeclareCaptionStyle{ruled}{labelfont=normalfont,labelsep=colon,strut=off} 
\floatstyle{ruled}
\newfloat{listing}{tb}{lst}{}
\floatname{listing}{Listing}

\usepackage{booktabs}
\usepackage[table]{xcolor}
\definecolor{SectionGray}{gray}{0.88}
\definecolor{GainGreen}{RGB}{0,128,96}
\newcommand{\gain}[1]{\makebox[0pt][l]{\textsuperscript{\scriptsize\textcolor{GainGreen}{#1}}}}

\title{Self-Play Meets Skill Evolution:\\Self-Evolving Search Agents that Pose, Solve, and Remember}
\author{
    Zenghuang Fu\textsuperscript{\rm 1,\rm 2}\equalcontrib,
    Zhaoyang Li\textsuperscript{\rm 3}\equalcontrib,
    Qiuyuan Ai\textsuperscript{\rm 3}\equalcontrib,
    Haoyu Wu\textsuperscript{\rm 3},\\
    Minghui Wu\textsuperscript{\rm 4},
    Chenxu Zhao\textsuperscript{\rm 4},
    Ante Wang\textsuperscript{\rm 5},
    Guannan He\textsuperscript{\rm 3}\corresponding,
    Changwei Wang\textsuperscript{\rm 6,\rm 7}\corresponding
}
\affiliations{
    \textsuperscript{\rm 1}University of Chinese Academy of Sciences\\
    \textsuperscript{\rm 2}Institute of Automation, Chinese Academy of Sciences\\
    \textsuperscript{\rm 3}Peking University\\
    \textsuperscript{\rm 4}Mininglamp Technology\\
    \textsuperscript{\rm 5}Tsinghua University\\
    \textsuperscript{\rm 6}Key Laboratory of Computing Power Network and Information Security,
    Ministry of Education; Shandong Computer Science Center, Qilu University of Technology
    (Shandong Academy of Sciences)\\
    \textsuperscript{\rm 7}Key Laboratory of Computing Power Internet and Service Computing,
    Shandong Fundamental Research Center for Computer Science
}

\begin{document}

\maketitle

\begin{abstract}
Self-play agents can generate training problems without questions from target
benchmarks, but their curricula lack persistent state: failures affect gradients
yet do not explicitly shape future practice. External skill memories preserve
procedural experience but are typically learned from fixed task distributions.
We introduce \textbf{SESA} (Self-Evolving Skill-Augmented Agent), which makes
procedural memory an evolving state of tool-augmented search self-play. A
challenger poses problems, while a separately parameterized solver alone
retrieves skills. Informative failures are distilled into reusable skills and
written back to memory. The updated memory changes solver behavior and success,
which changes the challenger's reward and the distribution of future problems;
the resulting frontier produces new failures that rewrite memory. This
bidirectional loop makes task generation and skill memory co-evolve. Because
retrieved skills shape on-policy training trajectories, their benefits can
enter the model parameters as well as remain in the external bank, enabling
memory-free deployment and optional inference-time retrieval. Across seven
open-domain and multi-hop question-answering benchmarks, SESA improves average
accuracy over SSP by 1.2--3.2 points across multiple backbones and surpasses the
skill-augmented SkillRL baseline by 0.9 points under a unified evaluation
protocol. On Qwen3 models, SESA-Off retains 1.8--2.2 points of improvement over
SSP, while the final skill bank adds a further 0.5--1.0 points. These results
show that evolving skill memory is not merely an inference-time plug-in: it
changes policy learning and the future training distribution while retaining
value as optional external memory. Our code is available at
\url{https://github.com/Zenghuang-Fu/SESA-Self-Evolving-Search-Agents}.
\end{abstract}


\begin{figure}[!t]
\includegraphics[width=\linewidth]{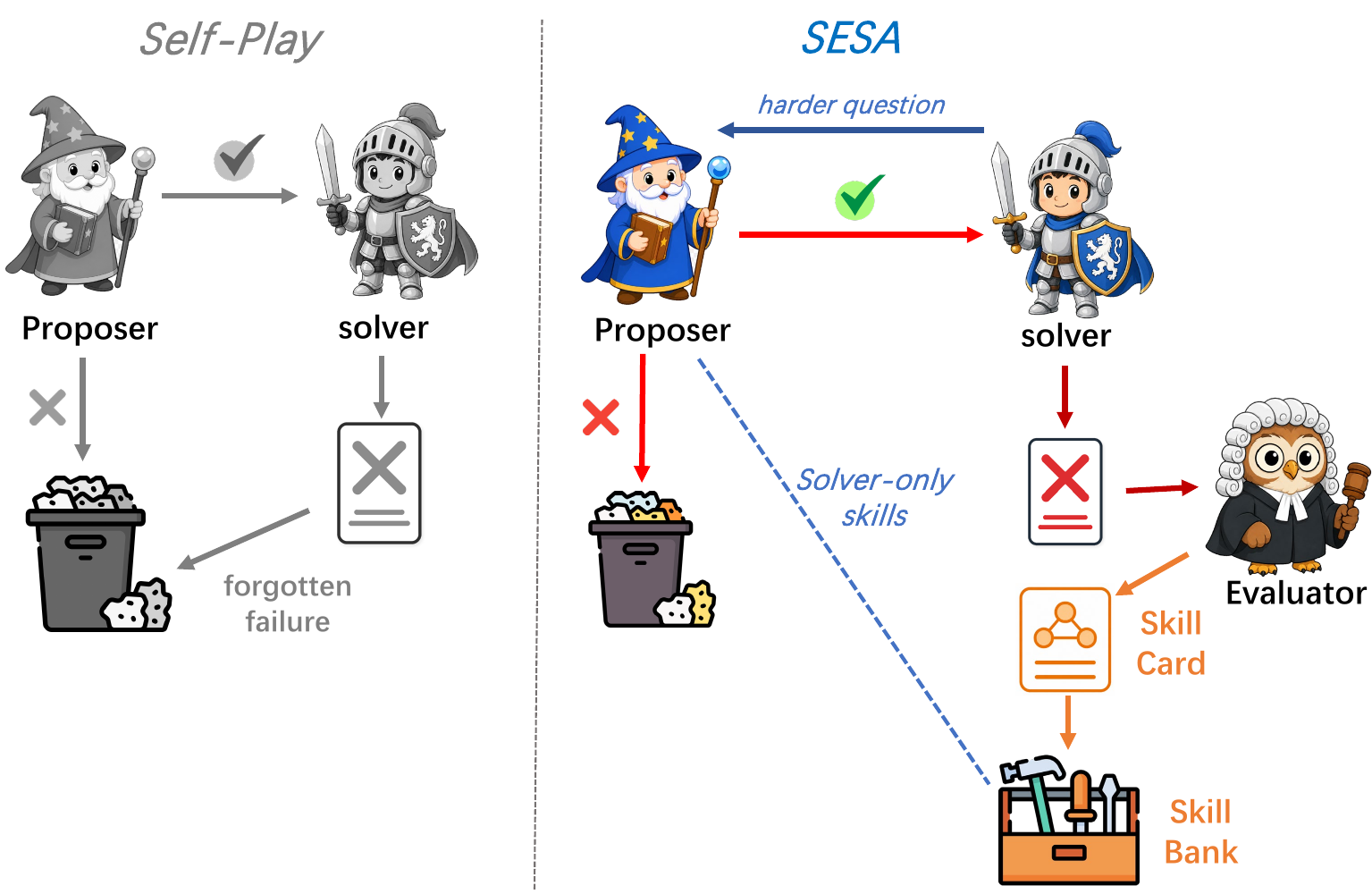}
\caption{Conceptual overview of SESA. Self-posed search failures are not discarded: they are distilled into reusable skills, stored in memory, and fed back to the solver so the next self-play round moves toward harder problems.}
\label{fig:figure1}
\end{figure}
\section{Introduction}

Large language model (LLM) agents are increasingly trained from their own
experience rather than a fixed corpus of demonstrations. One route is
\emph{zero-data self-play}, where an agent poses and solves problems using
verifiable rewards without an external question set
~\citep{lu2026search,chen2025multi,xia2025agent0,acikgoz2026tool}. Because task generation is
endogenous, self-play can adapt difficulty as the solver improves. Yet its
experience is usually transient: a trajectory contributes a policy gradient
but leaves no explicit, reusable account of the strategy learned from it.

A complementary route, \emph{skill-augmented reinforcement learning}, distills
experience into retrievable strategies or structured notes
~\citep{xia2026skillrl,wang2026reinforcement,shi2026skill1,li2026skillgraph,li2026arise,zhang2026coevoskills}. Such memories preserve
procedural knowledge beyond the update that produced it, but they are typically
built from fixed datasets or hand-designed curricula. These two limitations are
mirror images: self-play decides what to practice but forgets its lessons,
whereas skill-augmented RL remembers lessons from tasks it did not choose. The
missing setting is an agent that can do both.

We present \textbf{SESA} (Self-Evolving Skill-Augmented Agent), which places
online skill evolution inside tool-augmented search self-play. A challenger
poses a question with a verifiable target, and a separately parameterized
solver attempts it with a search tool. Informative solver failures are distilled
into human-readable skills, deduplicated, and written to a bounded
non-parametric memory. The solver retrieves these skills in later rounds, so
past failures change future on-policy trajectories and the data used for policy
optimization. Memory remains hidden from the challenger, preventing direct
skill leakage and preserving an asymmetric game between problem generation and
problem solving.

Simply composing self-play and a skill bank is not sufficient. Storing every
failure would accumulate noise and redundancy; exposing the same memory to the
challenger could leak solution strategies into generated questions; and adding
retrieval only after training would not change the self-play curriculum. A
closed loop must therefore decide which failures are learnable, who may access
their distilled lessons, and how those lessons return to on-policy training.

SESA realizes this feedback loop through four stages. \emph{Memory priming}
provides an initial retrieval substrate; \emph{asymmetric self-play} gives
solver-only access to skills; \emph{frontier shaping} steers the challenger
toward problems near the solver's current competence boundary; and
\emph{failure distillation}
converts useful failed rollouts into new skills. The result is a
failure-to-skill-to-solver loop: self-posed problems expose weaknesses, those
weaknesses become reusable guidance, and the strengthened solver pushes the
challenger toward a new frontier.

Because skills participate during training, SESA supports two forms of reuse.
Skill-conditioned rollouts can leave \emph{parametric carryover} in the trained
solver, allowing memory-free inference. The final bank can also remain enabled
for additional \emph{non-parametric augmentation}. We isolate these effects by
comparing SSP, SESA with memory disabled (SESA-Off), and the same trained SESA
solver with memory enabled (SESA-On). This distinction shows whether skill gains
reside in the policy, the external bank, or both.

We evaluate on 3,125 held-out questions spanning seven factual and multi-hop
search benchmarks. In the completed runs, SESA-On improves average accuracy
over SSP by 2.3 points on Qwen3-4B, 3.2 points on Qwen3-8B, and 1.2 points on
LLaMA-3.1-8B, while exceeding the corresponding base models by 10.9, 7.0, and
10.8 points. These gains across model scale and family indicate that persistent
skill evolution adds value beyond self-play alone; the controlled Off/On
comparison further tests how much of that value is parametric.

We make three contributions.
\begin{itemize}
\item \textbf{Coupled self-evolution.} We unite self-posed problem generation
with persistent skill consolidation, enabling an agent to choose its practice
frontier and retain lessons from its own failures.

\item \textbf{The SESA loop.} Solver-only retrieval, frontier shaping, and
online failure distillation feed reusable skills back into subsequent self-play
without leaking memory to the challenger.

\item \textbf{Dual-path evaluation.} We separate memory-free parametric
carryover from inference-time retrieval gains across multiple search
benchmarks and model families.
\end{itemize}

\section{Related Work}

\subsection{Self-Play for Agent Training}
Self-play has become a practical way to train agents without a fixed pool of
human-written tasks. Search Self-Play (SSP)~\citep{lu2026search} trains proposer and
solver policies for retrieval-augmented search using only a verifiable reward;
Multi-Agent Evolve~\citep{chen2025multi} extends this idea with a
proposer--solver--judge game; Tool-R0~\citep{acikgoz2026tool} studies zero-data
self-play for tool use; and EvolveR~\citep{wu2025evolver} frames self-evolution as an
experience-driven lifecycle. R-Few~\citep{yu2025guided} and Agent0~\citep{xia2025agent0}
further emphasize asymmetric roles and adaptive difficulty. These methods make
task generation endogenous, but the solving experience is usually consumed as a
training trajectory and then discarded. SESA follows the self-posed setting but
adds an explicit consolidation path: failed solver rollouts become retrievable
skills that affect later self-play rounds.

\subsection{Skill Memory and Experience Consolidation}
A complementary line studies agents that store reusable experience outside the
model weights~\citep{ai2026cognitive}. SkillRL~\citep{xia2026skillrl} combines a cold-start skill bank,
retained failures, and policy--skill co-evolution; related work further studies
reinforcement learning with skill libraries~\citep{wang2026reinforcement}, as well as skill
rollout, selection, structure, and co-evolution in Skill1~\citep{shi2026skill1},
SkillGraph~\citep{li2026skillgraph}, ARISE~\citep{li2026arise}, and
CoEvoSkills~\citep{zhang2026coevoskills}. Other systems focus on skill curation
and lifecycle management~\citep{ouyang2026skillos,pu2026skillops,lin2026muse,lin2026skillc}, while
Voyager~\citep{wang2024voyager}, Reflexion~\citep{shinn2023reflexion}, and ExpeL~\citep{zhao2024expel}
show that non-parametric memories can make agent experience reusable and
inspectable. However, these systems typically learn skills from fixed datasets,
hand-designed curricula, or non-RL interaction loops. SESA differs by placing
skill consolidation inside zero-data self-play: the agent both creates the
search problems that expose failures and writes those failures back into a
memory that changes future solving behavior. The distinction from SkillRL is
therefore not merely whether a skill bank is present. SkillRL evolves skills
under an exogenous task distribution, whereas SESA lets the solver's evolving
memory change its behavior on an endogenous frontier, which in turn changes the
reward that trains the challenger. Task generation and procedural memory thus
become coupled parts of the same learning process.

\section{Method}

\subsection{Setup and Notation}
\begin{figure*}[t!]
\centering
\includegraphics[width=\textwidth]{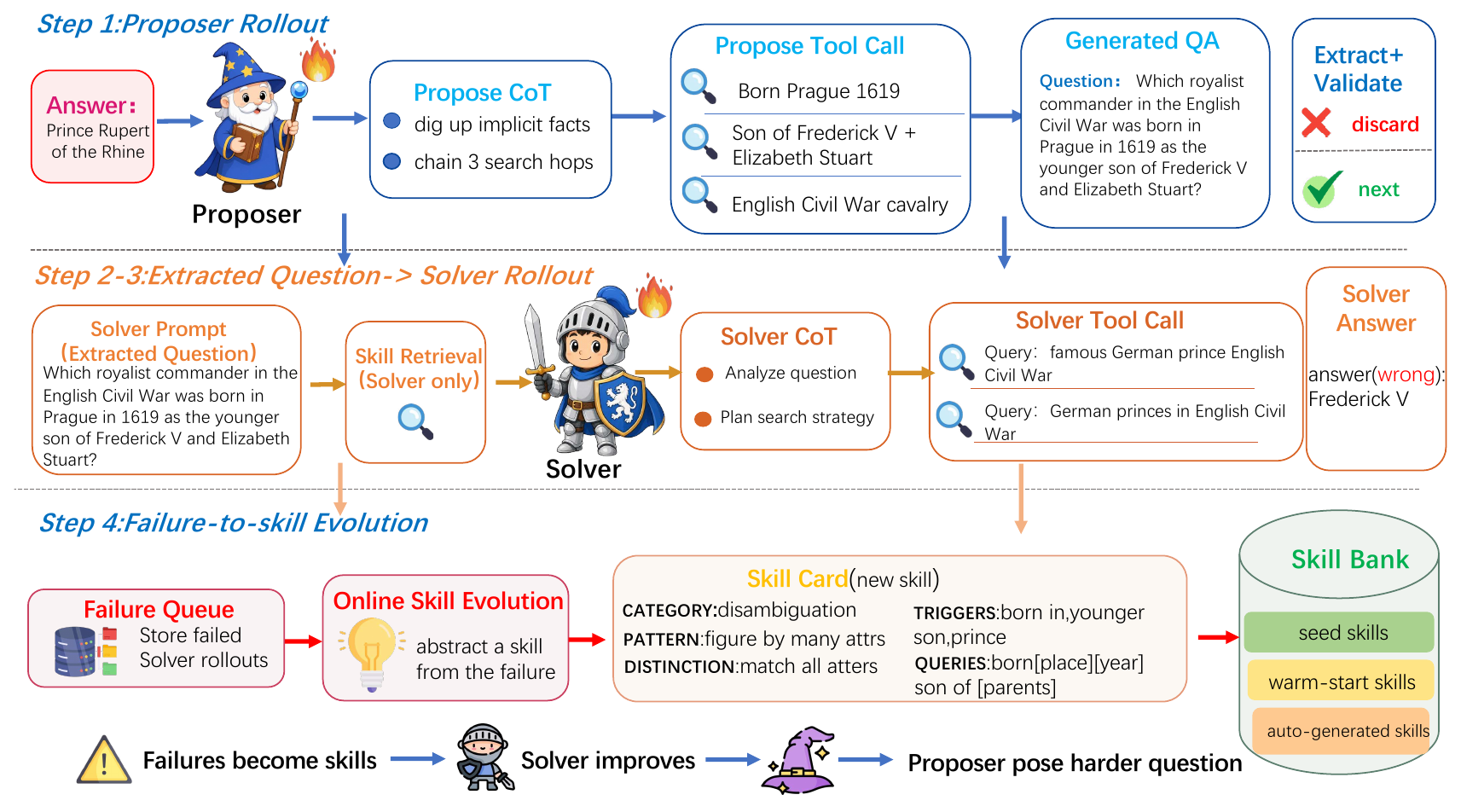}
\caption{The SESA training loop. Memory priming seeds a retrievable
skill bank; asymmetric self-play lets a challenger pose search tasks while only
the solver can retrieve skills; frontier shaping steers the challenger toward
problems near the solver's solvability boundary; and failure distillation converts
failed rollouts into deduplicated skills that are written back to memory. The
updated bank strengthens the solver and raises the frontier for subsequent
challenger-generated problems.}
\label{fig:main}
\end{figure*}

SESA trains a tool-augmented search agent through self-play, with no external
question set. A \emph{proposer} (challenger) policy $\pi_p$ generates a search
problem; a \emph{solver} (learner) policy $\pi_s$ attempts it by issuing
retrieval queries to a fixed search tool and producing a final answer; and a
verifiable reward compares the answer against the proposer-provided target. On
top of this self-play backbone, SESA maintains a non-parametric \emph{skill
memory} $\mathcal{B}$: a set of retrievable, human-readable strategies that the
solver consults during training and can optionally retain at inference time,
and that grows from the solver's own failures. Each skill is stored as
\begin{equation}
    s=(u,c,a,z,m),
\end{equation}
where $u$ is its description, $c$ the trigger conditions, $a$ avoidance cues
(anti-patterns or common confusions), $z$ reusable query templates, and $m$
the retrieval, helpfulness, and hurt counts used for maintenance. The training loop is organized into
four algorithmic stages:
\emph{memory priming}, \emph{asymmetric self-play},
\emph{frontier shaping}, and \emph{failure distillation}.
Figure~\ref{fig:main} shows how these stages close the SESA flywheel. We first
define the agentic reinforcement-learning objective that drives the self-play
game, then describe each stage below and analyze why they must couple in the
next section.

\subsection{Memory Priming}

SESA initializes $\mathcal{B}$ with 15 hand-written skills covering recurring
search patterns and 142 deduplicated skills mined during an earlier self-play
bootstrap:
\begin{equation}
    \mathcal{B}_0=\mathcal{B}_{\mathrm{seed}}\cup
    \mathcal{B}_{\mathrm{warm}}.
\end{equation}
Across the 157 initial entries, descriptions, triggers, avoidance cues, query
templates, and usage metadata provide an initial retrieval substrate and anchor
the granularity of later skill distillation and deduplication.

\subsection{Asymmetric Self-Play}

SESA uses separately parameterized proposer and solver policies, following
asymmetric self-play~\citep{xia2025agent0}. This separation gives the solver's success
rate a stable interpretation as the difficulty of proposer-generated problems
and, crucially, lets SESA expose retrieved skills only to the solver. The
proposer therefore adapts through reward feedback without directly observing
solution-oriented memory. Additional motivation and implementation details for
this information asymmetry appear in the supplementary material.

\subsection{Agentic RL Objective}

SESA optimizes both roles with a critic-free policy-gradient backend built on
Group Relative Policy Optimization (GRPO). The solver uses the standard grouped
form: for each generated problem $x=(q,a^\ast)$, it samples $G$ independent
search rollouts
\begin{equation}
    \tau_i \sim \pi_s(\cdot \mid q, \mathcal{T}, R(q;\mathcal{B}_t)),
    \quad i=1,\ldots,G,
\end{equation}
where $\mathcal{T}$ is the search tool and $R(q;\mathcal{B}_t)$ is the
solver-only retrieved skill context. Each rollout produces a final answer
\mbox{$\hat{a}_i$}. The solver receives a verifiable answer reward,
\begin{equation}
    r_s(\tau_i,a^\ast) =
    \mathbf{1}\{\mathrm{Judge}(\hat{a}_i,a^\ast)=1\},
\end{equation}
where the judge first checks exact match after normalization and otherwise uses
a model-based semantic match against the target answer. This reward is sparse
but reliable, and is assigned to the terminal response token.
Because the retrieved context $R(q;\mathcal{B}_t)$ enters the \emph{on-policy}
rollouts in Eq.~(1), skills do not merely condition a single inference pass:
they reshape the trajectory distribution on which the policy gradient is
computed. Retrieved guidance is thus internalized into the solver parameters
during training rather than acting only as an inference-time prompt, which is
why the trained solver retains most of its advantage even with the bank disabled
(cf.\ Table~\ref{tab:ablation-dualpath}).

The proposer is optimized against a difficulty-shaped reward derived from the
same solver group (defined in the next section), so that it learns to pose
problems near the solver's competence frontier. For solver updates, GRPO
normalizes rewards within the $G$ rollouts of the same problem,
\begin{equation}
    \hat{A}_i =
    \frac{R_i-\mathrm{mean}(\{R_j\}_{j=1}^{G})}
    {\mathrm{std}(\{R_j\}_{j=1}^{G})+\epsilon},
\end{equation}
and applies the resulting advantage to the generated response tokens. The
proposer emits one candidate per training instance and obtains its learning
signal only after the solver group has been evaluated, and is updated with the
same critic-free policy-gradient backend.

\subsection{Frontier Shaping}

Not every self-posed problem yields a useful gradient, and not every failure
yields a useful skill. Problems the solver always gets right carry no learning
signal; problems it always gets wrong are usually noise---outside the solver's
current reach---and the failures they produce, if consolidated, would pollute
the skill memory with un-actionable entries. SESA therefore keeps learning on
problems near the solver's competence boundary, using the same empirical
difficulty signal that the solver group already provides,
\begin{equation}
    \hat{p}_s(x)=\frac{1}{G}\sum_{i=1}^{G}
    \mathbf{1}\{r_s(\tau_i,a^\ast)>0\},
\end{equation}
the fraction of the $G$ solver rollouts that succeed on problem $x$.

Rather than hard-filtering batches, SESA shapes the proposer's reward so that
the challenger is trained to \emph{generate} frontier-difficulty problems in the
first place. A na\"ive complement-of-accuracy reward $1-\hat{p}_s(x)$ encourages
harder questions but over-rewards unsolvable ones and destabilizes self-play.
SESA instead uses a bell-shaped, endpoint-penalized reward,
\begin{equation}
    r_p(x)=
    \left\{
    \begin{array}{ll}
    -\lambda, & \hat{p}_s(x)\in\{0,1\},\\
    4(\ell+\hat{p}_s(x))(h-\hat{p}_s(x)), & \mathrm{otherwise},
    \end{array}
    \right.
\end{equation}
with $\ell=0$, $h=1$, and endpoint penalty $\lambda>0$. The reward peaks at
intermediate success rates and penalizes both trivial ($\hat{p}_s=1$) and
unsolvable ($\hat{p}_s=0$) questions, so the proposer is continually pushed to
pose problems just beyond the solver's current ability. As the solver improves,
the same reward drives the challenger toward a new frontier. Failures sampled
from this shaped distribution then flow into skill distillation, ensuring the
memory receives exploitable rather than noisy signal.

\subsection{Failure Distillation}

Failure distillation turns transient failures into durable, retrievable
knowledge; it is what the self-play line lacks. It runs as a three-phase
lifecycle synchronized with each training step.

\paragraph{Retrieval.}
Before the solver attempts a problem, SESA embeds the question and retrieves the
top-$k$ most similar skills from $\mathcal{B}$ using a dense encoder,
\begin{equation}
    R(q;\mathcal{B}_t) = \mathrm{TopK}_{s \in \mathcal{B}_t}
    \mathrm{sim}(e(q), e(s)),
\end{equation}
and prepends them to the solver's prompt as reference strategies. Retrieval is
read-only and deterministic given the current bank, and---by asymmetric
self-play---happens for the solver only.

\paragraph{Failure collection.}
After reward computation, failed solver rollouts are summarized into compact
records containing the problem, target, retrieved evidence, prediction, and
retrieved skill identifiers:
\begin{equation}
    \mathcal{F}_t=\{(q,a^\ast,\hat{a},R(q;\mathcal{B}_t))
    \mid r(\hat{a},a^\ast)=0\}.
\end{equation}
Only informative frontier failures enter a 300-record pending queue. Every 10
steps, once at least 20 have accumulated, consolidation selects at most 30,
prioritizing repeated failures and those unsolved despite retrieved guidance.

\paragraph{Consolidation.}
A judge abstracts each selected failure into its trigger, distinguishing
evidence, avoidance cues, and query templates, targeting what prior guidance
missed. A candidate is admitted only if its E5-base-v2 cosine similarity is at
most 0.93 against the bank and candidates already admitted in the same update:
\begin{equation}
    \max_{s'\in\mathcal{B}_t\cup\Delta\mathcal{B}^{<s}_t}
    \mathrm{sim}(e(s),e(s'))\leq 0.93.
\end{equation}
After admission, the maintained bank becomes
\begin{equation}
    \mathcal{B}_{t+1}=\mathrm{Maintain}
    (\mathcal{B}_t\cup\Delta\mathcal{B}_t).
\end{equation}
Seed skills are retained. A non-seed skill is evicted after at least three
retrievals if its helpful count minus hurt count is negative; overflow beyond
800 entries removes the lowest-scoring non-seed skills. At eligible step
boundaries, the trainer launches at most one consolidation job asynchronously.
Completion persists and increments the bank version seen by later retrievals.
Utility is assigned from the same solver rollouts: a correct answer increments
helpfulness for each retrieved skill, whereas a substantive incorrect answer
increments hurt; malformed trajectories are ignored. Retention therefore
reflects observed downstream behavior rather than age alone, and step-boundary
commits prevent the bank from changing within the solver batch that produced
the evidence.

Over training, this lifecycle makes the skill memory a living object: it grows
where the solver fails, forgets what does not help, and---because the failures
come from self-posed problems that get harder as the solver improves---keeps
acquiring skills for a difficulty frontier that no fixed dataset defines.

\section{Dual-Path Skill Reuse}

Because retrieved skills shape the on-policy trajectories used for training,
SESA can transfer experience through both model parameters and external memory.
Let $\theta_T$ and $\mathcal{B}_T$ denote the trained solver and final bank.
Memory-free deployment (\textsc{SESA-Off}) uses
\begin{equation}
    \pi_{\mathrm{off}}(\tau\mid q)
    =\pi_{\theta_T}(\tau\mid q,\mathcal{T}),
\end{equation}
whereas memory-augmented deployment (\textsc{SESA-On}) uses
\begin{equation}
    \pi_{\mathrm{on}}(\tau\mid q)
    =\pi_{\theta_T}(\tau\mid q,\mathcal{T},R(q;\mathcal{B}_T)).
\end{equation}
Comparing SSP with \textsc{SESA-Off} measures parametric carryover; comparing
\textsc{SESA-Off} with \textsc{SESA-On} isolates inference-time retrieval.
Because SESA has no explicit skill-distillation loss, carryover is an empirical
outcome rather than an architectural assumption. Further discussion is
provided in the supplementary material.

\section{Experiments}

\subsection{Experimental Setup}

\paragraph{Training data.}
SESA follows the zero-data setting of SSP: training does not consume questions
from any evaluation benchmark. We use the released SSP pool of 50,000 target
answers paired with one-, two-, or three-hop requirements (16,547/16,729/16,724
seeds). Preprocessing inserts each pair into the challenger prompt without
revealing the target. At every
iteration, the challenger must use the search engine to turn a sampled seed
into a concise, uniquely answerable question whose solution requires the
specified number of hops. The solver then attempts the generated question, and
informative failures produced under the frontier-shaped objective enter the
skill-distillation queue.
This separation ensures that improvements on the test benchmarks reflect
transfer from self-generated search experience rather than supervised exposure
to their questions.

\begin{table*}[!t]
\centering
\footnotesize
\setlength{\tabcolsep}{3.5pt}
\begin{tabular}{lcccccccc}
\toprule
\textbf{Method} & \textbf{NQ} & \textbf{TriviaQA} & \textbf{PopQA} &
\textbf{HotpotQA} & \textbf{2Wiki} & \textbf{MuSiQue} &
\textbf{Bamboogle} & \textbf{Avg.} \\
\midrule
\rowcolor{SectionGray}
\multicolumn{9}{c}{\textbf{Continual Training on Qwen3 Backbones}} \\
\midrule
Qwen3-4B & 46.4 & 65.8 & 45.0 & 42.8 & 43.0 & 20.0 & 54.4 & 45.3 \\
\quad + SSP & 54.4 & 76.8 & 53.6 & 56.4 & \textbf{52.8} & 25.4 & 57.6 & 53.9 \\
\quad + SESA & \textbf{56.2}\gain{+9.8} & \textbf{80.4}\gain{+14.6} & \textbf{55.2}\gain{+10.2} & \textbf{57.8}\gain{+15.0} & 51.8\gain{+8.8} & \textbf{27.2}\gain{+7.2} & \textbf{64.8}\gain{+10.4} & \textbf{56.2}\gain{+10.9} \\
Qwen3-4B-Instruct & 48.8 & 71.8 & 42.6 & 52.0 & 35.6 & 21.8 & 50.4 & 46.1 \\
\quad + SSP & 57.6 & 75.4 & 51.6 & \textbf{59.2} & 49.8 & \textbf{28.6} & \textbf{64.0} & 55.2 \\
\quad + SESA & \textbf{63.6}\gain{+14.8} & \textbf{80.4}\gain{+8.6} & \textbf{56.0}\gain{+13.4} & \textbf{59.2}\gain{+7.2} & \textbf{54.0}\gain{+18.4} & \textbf{28.6}\gain{+6.8} & 63.2\gain{+12.8} & \textbf{57.9}\gain{+11.8} \\
Qwen3-8B & 53.6 & 76.0 & 50.8 & 54.2 & 48.0 & 26.6 & 58.4 & 52.5 \\
\quad + SSP & 56.0 & 78.2 & 55.0 & 58.0 & 51.5 & 28.0 & \textbf{67.2} & 56.3 \\
\quad + SESA & \textbf{62.2}\gain{+8.6} & \textbf{82.8}\gain{+6.8} & \textbf{57.0}\gain{+6.2} & \textbf{64.0}\gain{+9.8} & \textbf{54.6}\gain{+6.6} & \textbf{32.6}\gain{+6.0} & 63.2\gain{+4.8} & \textbf{59.5}\gain{+7.0} \\
\midrule
\rowcolor{SectionGray}
\multicolumn{9}{c}{\textbf{Continual Training on Qwen2.5 Backbones}} \\
\midrule
Qwen2.5-7B-Base & 32.0 & 33.2 & 25.0 & 18.0 & 10.8 & 11.0 & 26.4 & 22.3 \\
\quad + SSP & 54.2 & \textbf{73.6} & 56.0 & 52.8 & 33.2 & 24.0 & \textbf{47.2} & 48.7 \\
\quad + SESA & \textbf{58.8}\gain{+26.8} & 72.2\gain{+39.0} & \textbf{61.4}\gain{+36.4} & \textbf{53.8}\gain{+35.8} & \textbf{38.0}\gain{+27.2} & \textbf{26.0}\gain{+15.0} & 45.6\gain{+19.2} & \textbf{50.8}\gain{+28.5} \\
Qwen2.5-7B-Instruct & 44.2 & 64.0 & 36.4 & 45.0 & 32.8 & 16.8 & 51.2 & 41.5 \\
\quad + SSP & 54.8 & \textbf{73.4} & 51.8 & 51.8 & 38.8 & 21.2 & \textbf{54.4} & 49.5 \\
\quad + SESA & \textbf{57.4}\gain{+13.2} & 72.2\gain{+8.2} & \textbf{55.2}\gain{+18.8} & \textbf{52.0}\gain{+7.0} & \textbf{42.2}\gain{+9.4} & \textbf{27.0}\gain{+10.2} & 51.2\gain{+0.0} & \textbf{51.0}\gain{+9.5} \\
\midrule
\rowcolor{SectionGray}
\multicolumn{9}{c}{\textbf{Continual Training on Cross-Family Backbones}} \\
\midrule
LLaMA-3.1-8B & 50.2 & 65.2 & 45.8 & 34.6 & 19.4 & 11.4 & 30.4 & 36.7 \\
\quad + SSP & 58.0 & 75.8 & \textbf{55.4} & 44.2 & 34.4 & \textbf{16.2} & \textbf{40.0} & 46.3 \\
\quad + SESA & \textbf{61.2}\gain{+11.0} & \textbf{79.2}\gain{+14.0} & 55.2\gain{+9.4} & \textbf{47.0}\gain{+12.4} & \textbf{35.2}\gain{+15.8} & 15.2\gain{+3.8} & 39.2\gain{+8.8} & \textbf{47.5}\gain{+10.8} \\
\midrule
\rowcolor{SectionGray}
\multicolumn{9}{c}{\textbf{Continual Training on Search-Specialized Agents}} \\
\midrule
Search-R1-7B & 56.6 & 75.4 & 57.2 & 58.2 & 45.2 & 29.6 & 55.2 & 53.9 \\
\quad + SSP & 57.8 & 78.0 & 58.4 & 60.4 & \textbf{45.6} & 30.6 & \textbf{59.2} & 55.7 \\
\quad + SESA & \textbf{63.0}\gain{+6.4} & \textbf{80.4}\gain{+5.0} & \textbf{60.8}\gain{+3.6} & \textbf{62.8}\gain{+4.6} & \textbf{45.6}\gain{+0.4} & \textbf{32.0}\gain{+2.4} & 57.6\gain{+2.4} & \textbf{57.5}\gain{+3.6} \\
\bottomrule
\end{tabular}
\caption{Answer accuracy (\%) on seven held-out search benchmarks. Results are
grouped by backbone family and search specialization; within each block,
\textit{+ SSP} and \textit{+ SESA} are initialized independently from the
corresponding unindented backbone.
Superscripts on SESA rows show absolute gains over the corresponding base
model. The best score within each backbone block is bold.}
\label{tab:main-results}
\end{table*}

\paragraph{Evaluation datasets.}
We evaluate on 3,125 held-out questions from seven benchmarks. Natural
Questions (NQ)~\citep{kwiatkowski2019natural}, TriviaQA~\citep{joshi2017triviaqa}, and
PopQA~\citep{mallen2023not} primarily test open-domain factual retrieval;
HotpotQA~\citep{yang2018hotpotqa}, 2WikiMultiHopQA (2Wiki)~\citep{ho2020constructing},
and MuSiQue~\citep{trivedi2022musique} emphasize compositional multi-hop search; and
Bamboogle~\citep{press2023measuring} provides a compact, challenging set of 125
questions that are difficult to answer without explicit decomposition. We use 500 examples
from each of the first six datasets and all 125 Bamboogle examples. This mix
tests whether the skills learned from self-posed problems transfer across both
fact-oriented and multi-hop distributions.

\paragraph{Models and baselines.}
We study Qwen3-4B, Qwen3-4B-Instruct, and Qwen3-8B~\citep{yang2025qwen3}; Qwen2.5-7B-Base and Qwen2.5-7B-Instruct~\citep{qwen2025qwen25technicalreport};
LLaMA-3.1-8B~\citep{grattafiori2024llama}; and the search-specialized
Search-R1-7B~\citep{jinsearch}. For each backbone, \emph{Base} denotes the
pretrained checkpoint before continual training, \emph{SSP} denotes self-posed
self-play~\citep{lu2026search} without a skill bank, and \emph{SESA} adds the
closed-loop skill evolution described in our method. Base, SSP, and SESA use
the same search backend and answer format. SSP and SESA are trained from the
same corresponding initialization; SESA differs only in the skill path unless
stated otherwise.

\paragraph{Metrics and evaluation protocol.}
The main metric is answer accuracy (\%), averaged equally over the seven
dataset-level scores. We first apply normalized exact match; predictions that
do not match lexically are checked for semantic equivalence by
Qwen2.5-32B-Instruct~\citep{qwen2025qwen25technicalreport}. We use greedy decoding with one rollout per question and at
most 10 assistant/search turns. Exact match and token-level F1 are retained as
diagnostic metrics but are not mixed into the main-table average. Unless
stated otherwise, the main table reports SESA-On.

\paragraph{Implementation details.}
We train with GRPO using five solver rollouts per generated problem. Retrieval
returns the top three E5-base-v2 records, and DeepSeek-v4-pro performs skill
distillation. Training uses $8\times$ NVIDIA A100-SXM4-80GB GPUs. Full
optimization, sequence-length, and distributed-training settings appear in the
supplementary material.

\subsection{Main Results}

Table~\ref{tab:main-results} compares each base search agent with SSP and SESA.
This layout separates the gain from self-play itself (Base to SSP) from the
additional gain of making self-play experience persistent (SSP to SESA), while
holding the backbone fixed.

\paragraph{Persistent skills improve self-play across scales.}
On the three Qwen3 backbones, SESA improves average accuracy over SSP by 2.3
points on Qwen3-4B, 2.7 points on Qwen3-4B-Instruct, and 3.2 points on Qwen3-8B.
The corresponding gains over the untrained checkpoints are 10.9, 11.8, and 7.0
points. The advantage therefore does not vanish as the backbone grows or after
instruction tuning: self-play provides the first improvement, while persistent
failure consolidation adds a further gain. The same trend appears on the two
Qwen2.5 settings, where SESA exceeds SSP by 2.1 and 1.5 average points.

\paragraph{The gain transfers across model families.}
On LLaMA-3.1-8B, SESA reaches 47.5 average accuracy, improving over SSP by 1.2
points and over the base model by 10.8 points. On the search-specialized
Search-R1-7B initialization, it reaches 57.5 and remains above SSP, showing that
the skill loop still contributes after search-oriented training. Improvements
are not uniform at the dataset level: Qwen3-4B is slightly below SSP on 2Wiki,
and Qwen3-8B is lower on Bamboogle. We therefore interpret SESA as a consistent
average improvement across initializations, rather than a guarantee of
monotonic gains on every benchmark.

\subsection{Where Do Skill Gains Reside?}

To separate improvements carried by the trained policy from improvements that
require retrieval at inference time, we evaluate three controlled modes:
an SSP-trained solver without memory; a SESA-trained solver with the skill bank
disabled (\textsc{SESA-Off}); and the same SESA-trained solver with the final
bank enabled (\textsc{SESA-On}). All three share the same search tool and
decoding configuration, and \textsc{SESA-Off} and \textsc{SESA-On} use identical
weights, so any gap between them comes purely from inference-time retrieval.
This design decomposes the benefit of skills into two additive paths:
\emph{parametric carryover} (\textsc{SESA-Off} over SSP), i.e.\ capability that
skill-conditioned self-play leaves inside the policy even after memory is
removed; and \emph{retrieval benefit} (\textsc{SESA-On} over \textsc{SESA-Off}),
i.e.\ the residual value of the external bank at test time.

Table~\ref{tab:ablation-dualpath} shows that the parametric path is consistently
positive. Relative to SSP, \textsc{SESA-Off} gains 1.8 points on Qwen3-4B and
2.2 points on Qwen3-8B, even though no skills are retrieved at evaluation time.
Re-enabling the same final bank adds another 0.5 and 1.0 points, respectively.
The dataset-level effect of retrieval is mixed: relevant guidance can help, but
irrelevant context can also distract the solver. Thus most of SESA's average
gain resides in the trained policy, while the bank is best viewed as an
optional, model- and task-dependent augmentation.

\begin{table}[!t]
\centering
\scriptsize
\setlength{\tabcolsep}{2.5pt}
\resizebox{\columnwidth}{!}{%
\begin{tabular}{lcccccccc}
\toprule
\textbf{Mode} & \textbf{NQ} & \textbf{TQA} & \textbf{PQA} &
\textbf{HQA} & \textbf{2Wi} & \textbf{MSQ} &
\textbf{BBL} & \textbf{Avg.} \\
\midrule
\rowcolor{SectionGray}
\multicolumn{9}{c}{\textbf{Qwen3-4B}} \\
\midrule
SSP (no mem.) & 54.4 & 76.8 & 53.6 & 56.4 & 52.8 & 25.4 & 57.6 & 53.9 \\
SESA-Off & \textbf{57.8} & 78.4 & \textbf{56.6} & \textbf{58.2} & \textbf{53.6} & 25.0 & 60.0 & 55.7 \\
SESA-On & 56.2 & \textbf{80.4} & 55.2 & 57.8 & 51.8 & \textbf{27.2} & \textbf{64.8} & \textbf{56.2} \\
\midrule
\rowcolor{SectionGray}
\multicolumn{9}{c}{\textbf{Qwen3-8B}} \\
\midrule
SSP (no mem.) & 56.0 & 78.2 & 55.0 & 58.0 & 51.5 & 28.0 & \textbf{67.2} & 56.3 \\
SESA-Off & 57.8 & 81.4 & \textbf{57.6} & 60.2 & \textbf{56.4} & 31.2 & 64.8 & 58.5 \\
SESA-On & \textbf{62.2} & \textbf{82.8} & 57.0 & \textbf{64.0} & 54.6 & \textbf{32.6} & 63.2 & \textbf{59.5} \\
\bottomrule
\end{tabular}%
}
\caption{\textbf{Dual-path ablation.} SSP uses no memory; SESA-Off and SESA-On
share identical weights and differ only in whether the skill bank is enabled at
inference. Best per column within each backbone is in bold.}
\label{tab:ablation-dualpath}
\end{table}

\begin{figure*}[!t]
\centering
\includegraphics[width=0.95\textwidth]{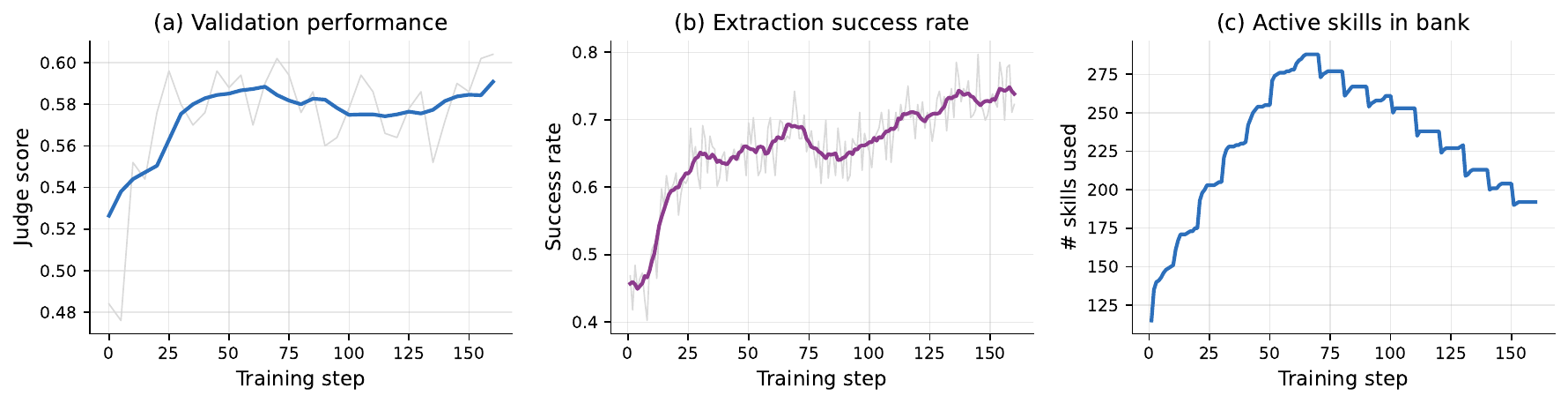}
\caption{\textbf{Self-evolution dynamics during training} (Qwen3-4B).
\textbf{(a)} Validation judge score rises early and plateaus at a high level.
\textbf{(b)} The challenger's problem-extraction success rate increases steadily,
indicating progressively more solvable, well-formed self-play problems.
\textbf{(c)} The count of active skills (retrieved at least once) expands as new
skills are distilled and then contracts under deduplication and negative-utility
eviction, reflecting a self-refining memory rather than unbounded growth. Light
lines are raw per-step values; solid lines are smoothed.}
\label{fig:dynamics}
\end{figure*}

This directly supports our second claim: skill-conditioned self-play is not a
prompt-time trick but a training mechanism, since its gains persist in the
policy under \textsc{SESA-Off}, while the retained bank remains available as an
optional, lightweight enhancement at deployment.

\subsection{Component Ablations}

Table~\ref{tab:ablation-components} uses leave-one-out retraining to test memory
priming, frontier shaping, and failure distillation. Solver-only access is
treated as a design constraint because the stored skills are procedural solving
guidance; the supplementary material discusses this choice in detail.

\begin{table}[!t]
\centering
\scriptsize
\setlength{\tabcolsep}{2.5pt}
\resizebox{\columnwidth}{!}{%
\begin{tabular}{lcccccccc}
\toprule
\textbf{Variant} & \textbf{NQ} & \textbf{TQA} & \textbf{PQA} &
\textbf{HQA} & \textbf{2Wi} & \textbf{MSQ} &
\textbf{BBL} & \textbf{Avg.} \\
\midrule
SESA (full) & 56.2 & 80.4 & 55.2 & 57.8 & 51.8 & 27.2 & 64.8 & 56.2 \\
\quad $-$ memory priming & 59.2 & 76.4 & 56.0 & 58.0 & 47.8 & 26.4 & 59.2 & 54.7 \\
\quad $-$ frontier shaping & 53.2 & 76.4 & 57.2 & 55.8 & 51.4 & 25.6 & 58.4 & 54.0 \\
\quad $-$ failure distillation & 56.4 & 75.8 & 57.2 & 55.4 & 50.0 & 23.8 & 56.0 & 53.5 \\
\bottomrule
\end{tabular}%
}
\caption{\textbf{Component leave-one-out} on Qwen3-4B. Each row removes one
component while keeping the rest fixed; a larger average drop indicates a more
critical component.}
\label{tab:ablation-components}
\end{table}

Removing every component lowers the overall average. Without memory priming,
performance drops from 56.2 to 54.7, indicating that an initial retrieval and
schema anchor remains useful even after online skill growth begins. Removing
frontier shaping produces a larger 2.2-point drop, consistent with the need
to concentrate learning and consolidation on solvable failures. The largest
decrease, 2.7 points, occurs without failure distillation, directly supporting
the central claim that persisting lessons adds value beyond self-play updates.
Individual datasets vary, but the aggregate ordering identifies failure
distillation as the most consequential component in this study.

\begin{table}[!t]
\centering
\scriptsize
\setlength{\tabcolsep}{2.5pt}
\resizebox{\columnwidth}{!}{%
\begin{tabular}{lcccccccc}
\toprule
\textbf{Method} & \textbf{NQ} & \textbf{TQA} & \textbf{PQA} &
\textbf{HQA} & \textbf{2Wi} & \textbf{MSQ} &
\textbf{BBL} & \textbf{Avg.} \\
\midrule
Qwen2.5-7B-Instruct & 44.2 & 64.0 & 36.4 & 45.0 & 32.8 & 16.8 & 51.2 & 41.5 \\
\quad + SSP & 54.8 & \textbf{73.4} & 51.8 & 51.8 & 38.8 & 21.2 & \textbf{54.4} & 49.5 \\
\midrule
SkillRL-Search-7B & 53.6 & 69.4 & 47.8 & \textbf{54.8} & \textbf{45.6} & \textbf{29.2} & 50.4 & 50.1 \\
\midrule
SESA (ours) & \textbf{57.4} & 72.2 & \textbf{55.2} & 52.0 & 42.2 & 27.0 & 51.2 & \textbf{51.0} \\
\bottomrule
\end{tabular}%
}
\caption{\textbf{Comparison with skill-augmented RL} on the Qwen2.5-7B family
under a unified evaluation protocol. The best score in each column is bold.}
\label{tab:skillrl}
\end{table}

\subsection{Training Dynamics}

Figure~\ref{fig:dynamics} shows that validation quality rises and stabilizes as
the challenger produces more usable problems. Meanwhile, active skills first
expand and then contract under deduplication and eviction, indicating selective
memory refinement rather than unbounded accumulation. Additional definitions
and per-step statistics are provided in the supplementary material.

\subsection{Evidence for Coupled Evolution}

Three complementary observations connect the final gains to the proposed
feedback loop. First, \textsc{SESA-Off} outperforming SSP shows that skill use
changes the policy learned during self-play rather than merely adding test-time
context. Second, the 2.7-point decrease without failure distillation shows that
an evolving bank contributes beyond the initial skills. Third, the dynamics
trace simultaneous changes in validation quality, usable problem generation,
and active memory. The dynamics alone are correlational, but together with the
controlled ablations they support the intended mechanism: self-generated
failures alter the skill bank, and the updated bank changes subsequent learning
trajectories.

\subsection{Comparison with Skill-Augmented RL}

Because SESA bridges self-play and skill evolution, SSP alone is not a
sufficient baseline. We additionally evaluate the released
SkillRL-Search-7B~\citep{xia2026skillrl} checkpoint under the same search backend,
decoding, and semantic-judging protocol. As shown in
Table~\ref{tab:skillrl}, SkillRL reaches 50.1 average accuracy, exceeding the
Qwen2.5-7B-Instruct SSP baseline by 0.6 points. SESA reaches 51.0 under the same
protocol and outperforms SkillRL by 0.9 points. This controlled comparison
indicates that coupling skill evolution to an endogenous task frontier provides
additional value over fixed-dataset skill learning.

\section{Conclusion}

SESA couples self-posed self-play with persistent skill evolution by distilling
frontier failures into a maintained memory that changes subsequent training.
Across model scales, families, and search-specialized initializations, this
closed loop consistently improves average accuracy over SSP, with component
ablations identifying online failure distillation as the largest contributor.
The Off/On evaluation further shows that skill-conditioned training leaves
substantial capability in the model parameters, while the retained bank
provides smaller, task-dependent inference gains. SESA thus treats procedural
memory as evolving training state rather than an inference-only prompt,
supporting both memory-free and memory-augmented deployment.

\bibliography{aaai2027}
\end{document}